# From Clay to Code: Typological and Material Reasoning in AI Interpretations of Iranian Pigeon Towers

Abolhassan Pishahang [1], Maryam Badiei[2]

[1] PhD Student, Comparative Studies (Art, Design, and Aesthetics Track), Florida Atlantic University, Boca Raton, United States , [2] PhD Student, Design Research, North Carolina State University, Raleigh, United States

apishahang2023@fau.edu; mbadiei@ncsu.edu

**Abstract.** This study investigates how generative AI systems interpret the architectural intelligence embedded in vernacular form. Using the Iranian pigeon tower as a case study, the research tests three diffusion models, Midjourney v6, DALL·E 3, and DreamStudio based on Stable Diffusion XL (SDXL), across three prompt stages: referential, adaptive, and speculative. A five-criteria evaluation framework assesses how each system reconstructs typology, materiality, environment, realism, and cultural specificity. Results show that AI reliably reproduces geometric patterns but misreads material and climatic reasoning. Reference imagery improves realism yet limits creativity, while freedom from reference generates inventive but culturally ambiguous outcomes. The findings define a boundary between visual resemblance and architectural reasoning, positioning computational vernacular reasoning as a framework for analyzing how AI perceives, distorts, and reimagines traditional design intelligence.

## 1 Introduction

Vernacular architecture carries environmental intelligence embedded in its material, spatial, and climatic adaptations. Iranian pigeon towers exemplify this knowledge system: cylindrical mudbrick structures designed to harvest fertilizer through controlled nesting, ventilation, and shading. Their form arises not from aesthetic choice but from iterative dialogue between craft, ecology, and material performance (Beazley, 1966; Bourgeois, 1983; Rapoport, 1969).

Generative AI offers new ways to visualize and reinterpret such traditions, yet its understanding of architecture remains largely visual. Diffusion-based models can reproduce recognizable geometry but rarely grasp the environmental logic or cultural context that produced it (Chen et al., 2025; Croce et al., 2023). When applied to heritage typologies, these systems generate compelling images that risk flattening localized intelligence into global aesthetic patterns (Tiribelli et al., 2024).

This study investigates how AI interprets vernacular material reasoning through the case of the Iranian pigeon tower. It examines how textual and visual inputs influence the fidelity and inventiveness of AI outputs, asking whether computational systems can engage not only with what architecture looks like, but how it works. Through a structured experimental workflow, the research evaluates three generative models—Midjourney v6, DALL·E 3, and DreamStudio (Stable Diffusion XL)—across referential, adaptive, and speculative prompts to test the boundaries between imitation and interpretation (Dahy, 2019).

## 2  Theoretical Background

### 2.1  Vernacular Architecture as Environmental Intelligence

Vernacular architecture emerges from continuous negotiation between material, climate, and use. Rather than following abstract design rules, it evolves through observation, adaptation, and repair, each generation refining what already works. In arid regions, where temperature, light, and dust impose strict limits, material knowledge becomes ecological strategy. Earthen construction stores and releases heat, filters air, and weathers into breathable surfaces that stabilize interior comfort (Bourgeois, 1983).

The Iranian pigeon tower (kaboutarkhaneh) exemplifies this environmental reasoning. Built from sun-dried mudbrick, its cylindrical body, rhythmic perforations, and ventilated turrets provide thousands of nesting cells while regulating airflow and daylight for guano production in adjacent fields (Beazley, 1966; Momeni & Shiri, 2022). Geometry and performance coincide: the wall's porosity is both structure and ventilation, and the tower's mass anchors thermal stability. What might appear decorative is in fact climatic instrumentation.

Across cultures, such structures reveal architecture as a feedback system where spatial form emerges from the interaction of matter, climate, and human adaptation. In this view, design operates as a responsive process rather than a fixed intention—an idea echoed in performance-oriented and material-driven approaches that treat building as a negotiation between natural behavior and human agency (Dahy, 2019; Hensel et al., 2012). Vernacular builders achieved this through intuition and craft, while contemporary designers pursue it through computational experimentation. Both approaches depend on cycles of adjustment between constraint and opportunity, bridging traditional ecological intelligence and digital fabrication.

### 2.2  Generative AI and the Challenge of Interpretation

Generative artificial intelligence enters this continuum as a visual interpreter. Diffusion models such as Midjourney, DALL·E 3, and DreamStudio (based on Stable Diffusion XL) synthesize images by translating textual and visual data into probabilistic associations. Their output can be visually convincing yet conceptually shallow: they reconstruct *what* a form looks like without understanding *why* it exists  (Leach, 2022). In heritage and architectural research, these models already assist in reconstructing damaged artifacts or visualizing historical typologies, producing high-fidelity images with minimal human input (Croce et al., 2023). Yet their success in resemblance often conceals interpretive limits.

When applied to architectural heritage, AI consistently reproduces recognizable geometry but overlooks the environmental reasoning that originally shaped it. Diffusion models can extrapolate missing surfaces or details with remarkable precision, yet they rarely register material behavior or climatic adaptation (Chen et al., 2025). This imbalance between visual fidelity and contextual understanding is reinforced by training datasets that merge imagery from diverse regions and eras, generating an aesthetic average that smooths cultural distinctions into a global visual norm (Croce et al., 2023). Without transparency in data provenance and ethical oversight, such systems risk perpetuating homogenizing and colonial patterns of representation, where local architectural intelligence is reduced to stylized exoticism. (Tiribelli et al., 2024)

This homogenizing tendency is evident when prompting an AI with "Iranian pigeon tower." The model often merges features from unrelated adobe or desert architectures into a generic Middle-Eastern silhouette. The result is seductive but epistemically thin, a simulation of context stripped of its climatic and social logic. AI here operates as an image historian without memory of use or matter. For design research, this limitation reframes the central question: can computational systems learn not only to mimic architectural appearance but also to reconstruct the reasoning that produced it?

## 2.3   Toward Computational Vernacular Reasoning

The concept of computational vernacular reasoning emerges from this tension between embodied and algorithmic intelligence. It refers to how generative AI systems attempt to interpret the ecological and material logic embedded in traditional architecture through prompt-based translation. Both vernacular builders and AI models learn through feedback, one tactile, the other statistical, but their modes of understanding differ profoundly. The human builder infers behavior from resistance and response; the algorithm predicts form from probability.

By structuring prompts around typological, environmental, and cultural cues, this study tests whether diffusion models can move beyond stylistic imitation toward interpretive reconstruction. The Iranian pigeon tower provides a precise field for this experiment: its geometry is consistent enough to benchmark fidelity, yet its environmental logic is complex enough to expose interpretive gaps (Momeni & Shiri, 2022).The objective is not to prove that AI can design vernacular architecture, but to reveal how it reads vernacular intelligence, what aspects it recognizes, what it distorts, and what it invents when guidance weakens.

This study builds on the notion of reciprocal design, extending it from fabrication to interpretation. Rather than shaping materials through computation, it examines how computation itself interprets material intelligence. In doing so, generative modeling becomes a site of translation between human intention and algorithmic inference—a dialogue that mirrors the adaptive reasoning found in vernacular craft (Dahy, 2019; Hensel et al., 2012). By testing how AI re-creates the pigeon tower under different descriptive and visual constraints, the research positions artificial intelligence within a continuous lineage of environmental reasoning that spans from embodied tradition to digital abstraction.

The following section translates this conceptual framework into a structured methodology for evaluating how diffusion models visualize architectural intelligence across referential, adaptive, and speculative conditions.

## 3   Methodology

This study operationalizes *computational vernacular reasoning* through an experimental workflow evaluating how generative AI interprets architectural intelligence in vernacular form. The process, shown in **Figure 1**, proceeds through four stages: (1) vernacular reference, (2) prompt framework, (3) AI image generation, and (4) evaluation—linking architectural knowledge with computational experimentation.

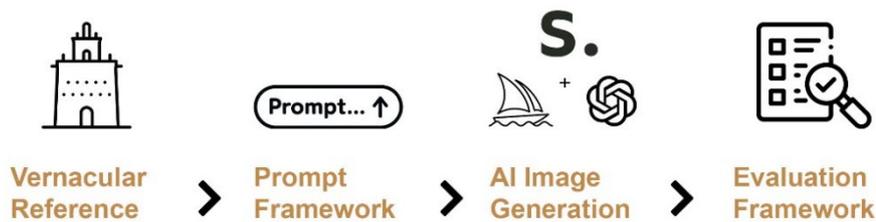

**Figure 1.** Methodological framework from vernacular reference to AI generation and evaluation across three prompt modes.

### 3.1   Vernacular Reference

The experiment focuses on the pigeon towers of Isfahan, Iran, chosen for their clear ecological performance and typological order. These cylindrical mudbrick dovecotes, pierced with rhythmic apertures and capped with ventilated turrets, historically provided nesting cells while harvesting guano for fertilizer (Amirkhani et al., 2010; Beazley, 1966). Their form integrates airflow, daylight, and thermal regulation into a unified environmental system (Momeni & Shiri, 2022). Morphological data from archival sources informed the textual prompts used in the experiment.

### 3.2   Prompt Framework

Three prompts were designed to test AI reasoning from reproduction to transformation:

**Prompt 1 – Referential:**
   Traditional Iranian pigeon tower constructed of sun-dried mudbrick, vertically cylindrical and clustered with a flat or slightly domed top and ventilated chimney caps. Rhythmic perforated openings for nesting pigeons arranged in horizontal bands along the façade. Built for collecting guano fertilizer, the structure shows thick adobe walls, interior honeycomb cells, and roof turrets for air exchange and bird entry. Set in an open desert landscape near Isfahan, showing handmade clay texture, weathered surface, and warm daylight tones. Realistic architectural photograph with balanced composition.– ar 3:4

**Prompt 2 – Adaptive:**
Adaptive reinterpretation of a traditional Iranian pigeon tower for present-day rural Isfahan. Cylindrical earthen structure redesigned to enhance airflow for nest hygiene and manure drying, with optimized aperture patterning and revised roof chimneys to manage bird access and ventilation. **Preserves** mudbrick texture and rhythmic perforations while integrating modular geometry for easier cleaning and fertilizer collection. Located in an arid agricultural field under warm desert daylight, photographed realistically to convey continuity between vernacular intelligence and modern environmental management. – ar 3:4

**Prompt 3 – Speculative:**
Speculative reimagining of the Iranian pigeon tower as a future ecological infrastructure. Inspired by the clustered geometry and material pragmatism of **Isfahan's** mudbrick towers, reinterpreted as a network of vertical habitats for birds and nutrient recycling. Forms respond to desert wind and light through porous skins and layered nesting cavities, combining large-scale guano harvesting with landscape integration. Presented in soft desert light with cinematic composition, expressing a visionary yet culturally grounded continuity between historical ingenuity and future sustainable agriculture. – ar 3:4

This sequence traces a continuum from replication to innovation, revealing how different levels of descriptive freedom affect typological and environmental fidelity..

### 3.3 AI Image Generation

Three diffusion-based systems represented distinct paradigms of generative design: **Midjourney v6** (aesthetic-perceptual), **DALL·E 3** (semantic-linguistic), and **DreamStudio / Stable Diffusion XL** (technical-transparent) (Derevyanko & Zalevska, 2023; Thampanichwat et al., 2025) Each received identical prompts under two conditions—with and without a reference image of the Ejgerd pigeon towers in Falavarjan, Isfahan (Photograph: Khodarahmi, 2005) to isolate the influence of visual grounding on interpretive fidelity.

### 3.4 Evaluation Framework

Outputs were assessed across five qualitative criteria derived from architectural representation and environmental performance studies: (1) Typology & Form, (2) Materiality, (3) Context & Environment, (4) Realism & Representation, and (5) Cultural Specificity.
Images were independently reviewed by the authors through **qualitative comparative analysis (QCA)** to ensure interpretive consistency. Uniform prompts and metrics minimized aesthetic bias, enabling differences to reflect AI interpretive behavior rather than stylistic variance..

## 4 Results

The comparison across the three diffusion models reveals a clear pattern: generative AI systems readily reproduce the geometric and visual order of vernacular architecture but only partially engage its material and environmental intelligence. Across the referential, adaptive, and speculative prompts, performance shifts from accurate replication toward imaginative reinterpretation, tracing a gradient from control to creativity. Reference imagery improved proportional and atmospheric realism yet consistently constrained variation, while the absence of visual grounding encouraged inventive but culturally ambiguous results. Together, the outcomes expose the threshold between recognition and reasoning in AI interpretation, where form is remembered, but environmental logic is only simulated.

### 4.1 Prompt 1 – Referential: Faithful reproduction of the Iranian pigeon tower

Across engines, the referential stage confirmed that AI systems recall geometry more reliably than material or environmental logic (Table 1). Midjourney v6 balanced typological fidelity with atmospheric realism, while DALL·E 3 privileged geometric precision over tactile depth. DreamStudio (SDXL) remained the least consistent, often merging Iranian and generic desert forms. When reference imagery was introduced, spatial proportion and contextual embedding improved, yet surface richness and irregularity diminished.

**Table 1.** Prompt 1 – Referential

|  | Midjourney v6 | DALL·E 3 (ChatGPT) | DreamStudio (SDXL) |
|---|---|---|---|
| no reference image | 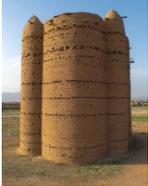 | 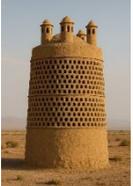 | 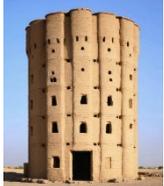 |
| Typology & Form | Coherent cylindrical tower; twin turrets; rhythmic apertures. | Precise cylindrical form; rigid aperture grid. | Simplified silo-like form; irregular openings. |
| Materiality | Believable earthen tone; mild smoothness. | Uniform clay surface; lacks tactile depth. | Flat polished texture; no shadowed detail. |
| Context & Environment | Realistic desert palette; photographic lighting. | Minimal background; isolated object. | Harsh light; weak environmental depth. |
| Realism & Representation | High photographic realism. | Balanced exposure; schematic tone. | Moderate realism; oversaturated color. |
| Cultural Specificity | Clearly vernacular; recognizably Iranian. | Generic adobe type; low specificity. | Blurred Middle Eastern reference; diluted identity. |
| with reference image 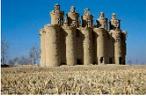 | 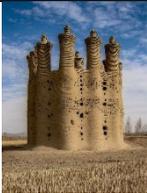 | 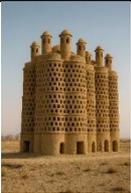 | 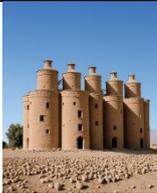 |
| **Typology & Form** | Enhanced proportions; sharper detail. | Expanded form; proportional accuracy. | Multi-tower cluster; hybrid typology. |
| **Materiality** | Improved clay realism; visible wear. | Slight tonal depth; minimal texture. | Convincing earthen tone; smooth finish. |
| **Context & Environment** | Integrated farmland; clear perspective. | Neutral landscape; flat horizon. | Realistic desert setting; credible depth. |
| **Realism & Representation** | Balanced atmospheric realism. | Technically crisp but static. | Cinematic tone; strong realism. |
| **Cultural Specificity** | Faithful to Iranian vernacular. | Accurate but generic. | Regionally plausible; mildly hybridized. |

The results suggest that descriptive prompts evoke structural accuracy, whereas visual references enforce formal coherence at the cost of vernacular imperfection. Together, they expose a recurring limitation: AI replicates the *appearance* of intelligence without reconstructing its environmental reasoning.

### 4.2 Prompt 2 – Adaptive: Testing contextual intelligence through controlled transformation

The adaptive stage examined how the three engines modified vernacular form when tasked with contextual reinterpretation rather than reproduction(Table 2). Midjourney v6 demonstrated the most coherent adaptation: maintaining typological stability while subtly refining proportion, ventilation rhythm, and environmental integration. DALL·E 3 retained structural accuracy but showed minimal transformation, its outputs appeared as refined replications rather than inventive redesigns. DreamStudio (SDXL) introduced environmental richness through vegetation, lighting, and social context but drifted toward stylistic hybridization, blending Iranian cues with generic desert imagery. Across models, reference imagery improved spatial grounding but further reduced creative deviation.

Overall, the adaptive results highlight a narrow interpretive bandwidth: AI systems can modify vernacular typologies aesthetically but struggle to translate their ecological rationale into functional design logic.

**Table 2**. Prompt 2 – Adaptive

| | Midjourney v6 | DALL·E 3 (ChatGPT) | DreamStudio (SDXL) |
|---|---|---|---|
| no reference image | 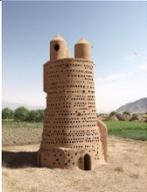 | 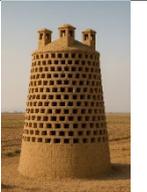 | 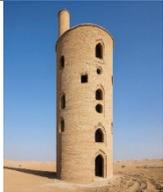 |
| Typology & Form | Balanced cylindrical form; subtle adaptation. | Geometrically precise; rigid rhythm. | Simplified shape; uneven apertures. |
| Materiality | Realistic adobe tone; slightly idealized. | Smooth clay surface; lacks texture. | Coarse finish; flat highlights. |
| Context & Environment | Believable mountain setting. | Sparse background; limited depth. | Harsh desert light; no spatial integration. |
| Realism & Representation | Photographic tone; strong coherence. | Balanced lighting; static rendering. | Realistic light; poor composition. |
| Cultural Specificity | Clearly Iranian; contextually credible. | Generic adobe type. | Ambiguous regional cues. |
| with reference image | 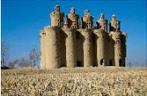 | | |
| Typology & Form | Stable geometry; refined details. | Accurate form; repetitive logic. | Multi-tower cluster; hybrid typology. |
| Materiality | Improved clay texture; still smooth. | Consistent tone; low variation. | Varied tone; enhanced depth. |
| Context & Environment | Integrated farmland; natural light. | Minimal change; neutral horizon. | Credible landscape; social context. |
| Realism & Representation | Balanced depth and realism. | Clear exposure; static scene. | Cinematic realism; lively tone. |
| Cultural Specificity | Faithful vernacular identity. | Accurate but generic. | Regionally plausible; stylistically mixed. |

## 4.3 Prompt 3 – Speculative: Probing the boundary between vernacular memory and creative synthesis

The speculative stage released the systems from typological constraint to test how they might extrapolate vernacular intelligence toward future-oriented forms (Table 3).

Without reference imagery, all three engines shifted from reproduction to invention: Midjourney v6 generated biomorphic clay towers that evoked natural growth; DALL·E 3 produced parametric lattice-like geometries rooted in recognizable logic; DreamStudio (SDXL) moved toward monumental, tectonic compositions with ambiguous function. These results demonstrate that abstraction enables AI to synthesize new morphologies, yet the link to environmental logic weakens as creative freedom increases. When the reference image was introduced, proportion and realism improved, but the imaginative range contracted, AI reverted to familiar typological safety. The stage thus underscores an inverse relationship between accuracy and creativity: the more the system recognizes the past, the less it invents beyond it.

**Table 3**. Prompt 3 – Speculative

| | Midjourney v6 | DALL·E 3 (ChatGPT) | DreamStudio (SDXL) |
|---|---|---|---|
| no reference image | 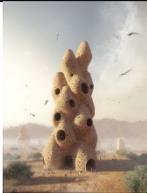 | 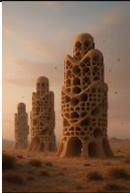 | 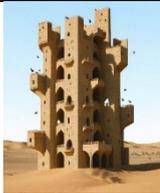 |
| Typology & Form | Organic sculptural tower; typology abstracted. | Ordered parametric lattice; rigid symmetry. | Fragmented angular mass; typology erased. |
| Materiality | Convincing earthen tone; tactile realism. | Smooth clay surface; digitally uniform. | Coarse shading; synthetic finish. |
| Context & Environment | Immersive desert light; strong atmosphere. | Warm desert tone; minimal context. | Flat desert scene; weak grounding. |
| Realism & Representation | Cinematic realism; atmospheric depth. | Balanced but static composition. | Realistic render; exaggerated geometry. |
| Cultural Specificity | Evokes vernacular memory through texture. | Generic desert futurism. | Culturally ambiguous; non-vernacular form. |
| with reference image 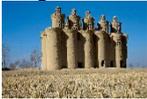 | 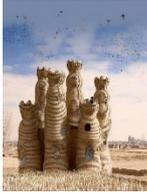 | 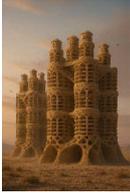 | 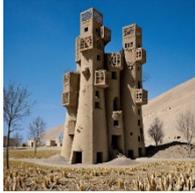 |
| Typology & Form | Multi-tower cluster; reduced novelty. | Layered stacks; proportional accuracy. | Hybrid fortress; coherent integration. |
| Materiality | Weathered clay tone; believable texture. | Enhanced shadow; still synthetic. | Rich tone range; solid surface. |
| Context & Environment | Natural horizon; lively desert field. | Soft dusk tone; neutral setting. | Rural backdrop; integrated terrain. |
| Realism & Representation | Photographic depth; detailed form. | High photorealism; limited concept. | Realistic lighting; bold geometry. |
| Cultural Specificity | Stylized vernacular echoes. | Globalized aesthetic. | Hybrid mix of regional and sci-fi cues. |

## 5 Discussion

The three prompt stages reveal a clear gradient from replication to invention, exposing how generative AI perceives vernacular intelligence as visual pattern rather than environmental reasoning. Across all engines, typological fidelity was consistent, AI could reproduce the cylindrical geometry, aperture rhythm, and massing of Iranian pigeon towers, but material and climatic logics remained abstracted.

Midjourney v6 achieved atmospheric coherence and credible context; DALL·E 3 excelled in geometric precision but lacked depth; DreamStudio (SDXL) offered compositional diversity at the expense of typological clarity.
The introduction of reference imagery consistently improved proportion, scale, and setting, yet reduced the expressive irregularity that defines vernacular craft. Without visual grounding, AI generated novel morphologies, particularly in the speculative stage, but detached them from climatic or material cause. This inverse correlation between realism and invention underscores a deeper limitation: diffusion models operate through pattern recognition, not environmental reasoning.
They reconstruct the look of adaptation, not the process that produced it.
Viewed through the lens of *computational vernacular reasoning*, these outcomes demonstrate both potential and boundary. AI can participate in the representational dialogue of vernacular form, but its interpretive power remains surface-bound, trained to emulate rather than to understand. Bridging this gap requires not more aesthetic tuning, but integration of environmental and material datasets that encode performance logic alongside visual description. Only then might generative systems move from image synthesis toward *architectural reasoning*, from recognition to reciprocity.

## 6      Conclusion

This study examined how generative AI systems interpret the typological, material, and environmental intelligence of Iranian pigeon towers through three levels of architectural prompting—referential, adaptive, and speculative.
Across all experiments, the diffusion models—Midjourney v6, DALL·E 3, and DreamStudio based on Stable Diffusion XL—proved adept at reproducing geometric form and surface realism but struggled to reconstruct the climatic and material reasoning that gives vernacular architecture its logic.
Reference imagery improved proportion and atmospheric accuracy yet limited creative exploration, while its absence encouraged invention at the cost of cultural specificity.
Evaluation across five criteria confirmed this imbalance: performance was strongest in Typology & Form and Realism & Representation, moderate in Context & Environment, and limited in Materiality—not in surface depiction, but in the absence of behavioral or climatic understanding of earthen construction.
Generative AI thus functions as a visual recognizer rather than a reasoning collaborator—it remembers but does not infer.
*Computational vernacular reasoning* is therefore proposed as a framework for assessing this interpretive boundary and guiding future integration of environmental and material intelligence into generative design.

**Acknowledgements.** The author acknowledges the use of ChatGPT (OpenAI, 2025) for editorial refinement and linguistic clarity, under full author supervision. Image generation and analysis were conducted using Midjourney v6, DALL·E 3 (OpenAI), and DreamStudio based on Stable Diffusion XL (Stability AI) as part of the experimental framework. All interpretations, datasets, and conclusions are the author's own.